%% file: acl_latex.tex
\pdfoutput=1

\documentclass[11pt]{article}

\usepackage[final]{acl}

\usepackage{times}
\usepackage{latexsym}
\usepackage{multirow}
\usepackage{booktabs}
\usepackage[T1]{fontenc}

\usepackage[utf8]{inputenc}
\usepackage{soul}
\usepackage{microtype}

\usepackage{inconsolata}

\usepackage{graphicx}

\title{CoPERLex: Content Planning with Event-based Representations for \\ Legal Case Summarization}

\author{Santosh T.Y.S.S, Youssef Farag, Matthias Grabmair \\ School of Computation, Information, and Technology; \\
Technical University of Munich, Germany \\ \ }

\begin{document}
\maketitle
\begin{abstract}
Legal professionals often struggle with lengthy judgments and require efficient summarization for quick comprehension. To address this challenge, we investigate the need for structured planning in legal case summarization, particularly through event-centric representations that reflect the narrative nature of legal case documents. We propose our framework, CoPERLex, which operates in three stages: first, it performs content selection to identify crucial information from the judgment; second, the selected content is utilized to generate intermediate plans through event-centric representations modeled as Subject-Verb-Object tuples; and finally, it generates coherent summaries based on both the content and the structured plan. Our experiments on four legal summarization datasets demonstrate the effectiveness of integrating content selection and planning components, highlighting the advantages of event-centric plans over traditional entity-centric approaches in the context of legal judgements.
\end{abstract}

\section{Introduction}
\input{text/introduction}

\section{Related Work}
\input{text/relatedwork}

\section{Our Method: CoPERLex}
\input{text/method}

\section{Experiments}
\input{text/experiments}

\section{Conclusion}
\input{text/conclusion}

\section*{Limitations}
\input{text/limitations}

\section*{Ethics Statement}
\input{text/ethics}

\bibliography{custom}

\appendix

\input{text/appendix}

\end{document}

%% file: text/introduction.tex
Legal professionals, including lawyers, judges, and researchers, often face the daunting task of navigating lengthy legal judgments that are critical for case law interpretation and reasoning. Automatic legal summarization can significantly aid this process by producing concise overviews of extensive documents \cite{hachey2006extractive}. Earlier efforts centered on extractive summarization, valued for its faithfulness to the original text \cite{farzindar2004atefeh,grover2003automatic,grover2003summarising,saravanan2006improving,bhattacharya2019comparative,polsley2016casesummarizer,liu2019extracting}. However, recent advancements have shifted towards abstractive approaches \cite{shukla2022legal,elaraby2023towards,elaraby2022arglegalsumm,moro2022semantic}, driven by the limitations of extractive methods, such as incomplete discourse, unresolved coreferences, and lack of context. Despite these, abstractive methods still face challenges in capturing the complex context of legal documents, often leading to hallucinations, highlighting the need for more accurate techniques \cite{deroy2023ready,deroy2024applicability}.

While most current approaches to legal case summarization rely on single end-to-end models to produce succinct, fluent, coherent, and factually accurate summaries, we argue that this setup often fails to effectively manage the implicit steps involved in summary generation, such as identifying salient content, organizing it logically, and maintaining faithfulness to the source. Moreover, this setup makes it difficult to mitigate specific types of errors without inadvertently introducing trade-offs that compromise other aspects of summary quality. These issues are particularly pronounced in the legal domain, where documents are often lengthy, dense, and rich in complex arguments and reasoning that require careful representation.

To address these limitations, we propose CoPERLex (Content Planning with Event-Based Representations for Legal Case Summarization), a three-stage framework inspired by how humans generate summaries from source documents. CoPERLex explicitly models the process as follows: (1) Content Selection, which involves analyzing the source document to identify the most salient content for inclusion in the summary; (2) Content Planning, where the selected content is organized into a structured intermediate representation, ensuring coherence and logical flow in the final summary; and (3) Summary Realization, which converts the intermediate representation into fluent, cohesive text.

For the \textbf{content selection} step, we employ an extractive summarization system, MemSum \cite{gu2022memsum}, which iteratively selects sentences based on a multi-step episodic Markov Decision Process, ensuring that the most relevant and salient content is identified for summarization.  In the subsequent \textbf{content planning} stage, the selected salient content is used to produce an intermediate summary representation, ensuring  coherence and logical structure for generating final summary. While various intermediate representations, such as entity chains \cite{narayan2021planning,adams2024speer,liu2021controllable,huot2024plan} or question-answer pairs as structural blueprints \cite{narayan2023conditional,huot2023text}, have been explored for short-text corpora, their application to lengthy legal judgments remains largely unexplored. Legal judgments, however, are inherently complex narratives that not only present facts but also describe a sequence of events, actions, and the legal reasoning applied to specific circumstances. This event-centric structure makes events crucial for understanding and summarizing such texts.  In light of this, we seek to leverage event representations as an intermediate representation to guide the legal summarization process. We adopt a simpler representation of events in the form of Subject-Verb-Object (SVO) tuples to capture key actions and events, beyond any jurisdiction- or case-specific taxonomy, as carried out in \citet{yao2022leven,shen2020hierarchical,li2020event}, for better generalizability. We extract event triples from the golden summaries to augment the dataset with target plans and train an encoder-decoder-based Unlimiformer model \cite{bertsch2023unlimiformer} to generate the intermediate representation based on the extracted salient content.

Finally, in the \textbf{summary realization} stage, the intermediate event-based plan, combined with the selected content, is fed into a Longformer Encoder-Decoder model \cite{beltagy2020longformer} to generate  summaries. Our experiments on four legal summarization datasets—MultiLexSumm-Long, MultiLexSumm-Short, CanLII, and SuperSCOTUS—demonstrate the effectiveness of CoPERLex. The integration of content selection and planning yields significant improvements in faithfulness and coherence, with event-centric plans outperforming entity-centric representations.

%% file: text/relatedwork.tex
\noindent \textbf{Legal Case Summarization:} Earlier works have predominantly utilized extractive approaches \cite{bhattacharya2019comparative} which encompass both unsupervised methods \cite{bhattacharya2021incorporating,polsley2016casesummarizer,farzindar2004atefeh,saravanan2006improving,Mandal2021ImprovingLC} and supervised methods \cite{agarwal2022extractive,liu2019extracting,zhong2019automatic,xu2021toward,Mandal2021ImprovingLC}. Unsupervised methods for legal summarization include domain-independent techniques, such as frequency-based statistics (e.g., Luhn \cite{luhn1958automatic}), graph-based methods (e.g., LexRank \cite{erkan2004lexrank}), and matrix factorization methods (e.g., LSA \cite{gong2001generic}). Additionally, legal-domain-specific approaches have been proposed, such as LetSum \cite{farzindar2004atefeh,saravanan2006improving}, which utilizes cue phrases to assign rhetorical and semantic roles to sentences in the source document and form a summary by selecting sentences, either through TF-IDF or K-mixture models.  \citet{polsley2016casesummarizer} use various domain-related features to score the importance of sentences, while \cite{zhong2019automatic} creates template-based summaries and employs Maximal Marginal Relevance to handle redundancy in selected sentences. \citet{bhattacharya2021incorporating} models summarization as an Integer Linear Programming optimization problem, incorporating guidelines from legal experts into the optimization setup. Additionally, \citet{Mandal2021ImprovingLC} utilizes document-specific catchphrases that provide both domain-specific and document-specific information to guide the summarization process. Supervised methods employ diverse strategies, ranging from the knowledge engineering of different domain-specific features \cite{liu2019extracting} to joint multi-task learning with Rhetorical Role Labeling \cite{agarwal2022extractive} and leveraging the argument structure within documents \cite{xu2021toward}. Recently, \citet{deroy2023ensemble} evaluated an ensemble of existing extractive-based supervised and unsupervised methods, employing simple voting-based techniques as well as ranking-based and graph centrality approaches.

With the limitations of extractive approaches in providing an overall coherent summary \cite{zhang2022extractive}, recent works explored abstractive summarization techniques for legal case summarization, especially considering the recent successes of the unsupervised pre-training and supervised fine-tuning paradigm \cite{shukla2022legal,ray2020summarisation,schraagen2022abstractive,santosh2024lexsumm,tyss2024lexabsumm}. \citet{shukla2022legal} applied various transformer-based pre-trained abstractive models, such as BART, Legal-LED, and Legal-Pegasus for legal case summarization. To address the challenge posed by long legal documents, \citet{moro2022semantic} proposed chunking input documents into semantically coherent segments, summarizing each chunk individually and then concatenating the results to avoid truncation. Recent efforts have also focused on enhancing the factuality of generated summaries \cite{feijo2023improving} by leveraging an entailment module to select faithful candidates. Additionally, rich argumentative structure embedded in legal documents are also explore to improve the performance of legal summarization models \cite{xu2022multi,elaraby2022arglegalsumm,elaraby2023towards}. While most of these works concentrate on training and evaluating summarization models within specific jurisdictions, recent research by \citet{santosh2024beyond} assesses the cross-jurisdiction generalizability of legal case summarization systems and further propose an adversarial learning approach to enhance cross-domain transfer of these methods.

To the best of our knowledge, no investigation has been conducted on planning-based approaches for legal summarization. In this work, we propose an event-centric representation as an intermediate plan for legal summarization.

\vspace{1em}

\noindent \textbf{Content Planning for Text Generation:}   Structured planning is considered a critical link in organizing content  effectively before realization \cite{reiter1997building}, much like humans plan at a higher level than individual words, as evidenced by psycholinguistic studies \cite{levelt1993speaking,guhe2020incremental}. Earlier approaches incorporated various planning representations, such as Rhetorical Structure Theory \cite{mann1988rhetorical,hovy1993automated} and MUC-style representations \cite{mckeown1995generating}, discourse trees \cite{mellish1998experiments}, entity transitions \cite{kibble2004optimizing,barzilay2008modeling}, sequences of propositions \cite{karamanis2004entity}, schemas \cite{mckeown1985discourse} and lexical chains \cite{barzilay1997using}. 

Recent works in the data-to-text generation task divide it into two distinct phases: planning and realization of natural language text \cite{moryossef2019step}. \citet{perez2018bootstrapping} introduced content selection methods that do not rely on explicit planning. Meanwhile, \citet{puduppully2019adata} and \citet{laha2020scalable} proposed micro-planning strategies, where they first establish a content plan based on a sequence of records and then generate a summary conditioned on that plan. Similar planning approaches have been explored for entity realization \cite{puduppully2019adata}. Additionally, \citet{puduppully2021data} advocated for macro-planning as a means of organizing high-level document content, either in text form or within latent space \cite{puduppully2022data}.

In summarization, various plan representations have been investigated.
\citet{narayan2020stepwise} treat step-by-step content selection as a planning component, generating sentence-level plans through extract-then-abstract methods \cite{zhang2022extractive}, and even sub-sentence-level plans using elementary discourse units \cite{adams2023generating}. Further research by \citet{narayan2021planning,liu2021controllable} introduced intermediate plans using entity chains—ordered sequences of entities mentioned in the summary. More recently, \citet{narayan2023conditional,huot2023text} conceptualized text plans as a sequence of question-answer pairs, which serve as blueprints for content selection (i.e., what to say) and planning (i.e., in what order) in summarization tasks. Despite these advancements, there remains no consensus on the most effective representation.

In this work, we explore the utility of content planning approaches for summarizing long legal documents, specifically by designing an event-based plan representation that aligns with the narrative structure inherent to these texts.

\noindent \textbf{Event-Centric Representations:} Understanding events is critical for various tasks, including open-domain question answering \cite{yang2003structured}, intent prediction \cite{rashkin2018event2mind}, timeline construction \cite{do2012joint}, text summarization \cite{daume2009bayesian}, and misinformation detection \cite{fung2021infosurgeon}. Events are not isolated predicates; they often interconnect through complex relationships, such as memberships, co-reference, temporal orders, and causal links. These relationships form narrative scripts, necessitating sophisticated representations that extend beyond simple predicates \cite{chen2021event}.

In the field of legal NLP, event-based representations have been underexplored for downstream tasks. However, recent work by \citet{joshi2023u} has utilized these representations for prior case retrieval, showcasing their potential. In this study, we leverage event-based representations to create an intermediate content plan specifically for legal case summarization. We adopt a simplified event representation from previous works \cite{chambers2008unsupervised,chambers2009unsupervised,jans2012skip}, where events are represented as verb-dependency pairs (e.g., subject, object), in contrast to jurisdiction- or case-specific taxonomy, as carried out in \citet{yao2022leven,shen2020hierarchical,li2020event}, for better generalizability.

%% file: text/method.tex
In this section, we present our proposed framework, CoPERLex, designed for the task of legal case summarization. Legal summarization requires processing complex and lengthy documents while ensuring that the generated summary is both coherent and faithful to the source. To address these challenges, we propose a pipelined approach that breaks down the task into three sub-tasks: (a) Content Selection Module identifies and selects the most salient sentences from the entire input document, similar to extractive summarization, ensuring that critical content is retained. (b) Content Planning Module takes the selected content as input and generates an intermediate content plan, serving as a blueprint for generating the final summary. This plan is constructed as event-based representations, expressed through subject-verb-object (SVO) triples. (c) Summary Generation Module synthesizes the summary by taking the selected content along with the event-based content plan as input, ensuring that the summary is structured and faithful to the outlined plan.

\subsection{Content Selection}
This module takes the input document segmented into sentences and outputs a set of salient sentences. We utilize the MemSum \cite{gu2022memsum} (Multi-step Episodic Markov decision process extractive SUMmarizer), a reinforcement-learning-based extractive summarizer that iteratively selects sentences for the summary. At each time step, the policy network (agent) based on the state decides either to stop or continue to extract next sentence from remaining list of unselected sentences. If the agent continues, it computes a score for each remaining sentence and samples a sentence based on the these scores.  The state is represented by (a) the local content of the sentence (b) the global context of the sentence within the document  and (c) the current extraction history (set of selected sentences till that point). To encode these three encodings in the state, we use a local sentence encoder, a global context encoder and an extraction history encoder. For local sentence encoder, we pass the word embeddings of each sentence through a bi-directional LSTM and multi-head pooling layer to obtain the sentence representation. The global context encoder takes each sentence embedding  from local encoder and passes them through a bi-LSTM to obtain context-based embedding of each sentence. Then extraction history encoder encodes the extraction history information and produces the extraction history embedding for each of the remaining unselected sentences through identical blocks, each containing two multi-head attention sublayers. The first sub-layer performs multi-head self-attention among the local embeddings of remaining sentences so that each remaining sentence can capture the context provided by other remaining sentences. The other attention sublayer performs multi-head attention over the embeddings of extracted sentences to enable each remaining sentence to attend to all the extracted sentences. This block finally outputs one history based embedding for each remaining sentence. Finally the local, global and extraction history based embedding of each unselected sentence is concatenated to produce score for each sentenece. We also multi-head pool all these embeddings to obtain the stopping probability. This step-wise state-updating strategy allows the agent to consider the content of the partial summary during sentence selection, contrasting with single-step episodic MDP-based models \cite{dong2018banditsum,narayan2018ranking}, which encode state information only once and extract the entire summary through a single action.

The policy network is trained using the Reinforce policy gradient algorithm \cite{williams1992simple}, with the objective of maximizing the expected return at each time step. This return is defined as the cumulative reward from time $t+1$ until the end of the episode, when the summary is complete. The instantaneous reward is zero, except at the end of the episode, where the final reward is computed as the mean of the ROUGE-1, 2, and L F1-scores between the extracted set of sentences (the summary) and the target abstractive summary. Notably, while training an extractive summarizer, we do not require explicit binary labels for each sentence as a supervision signal and instead utilize the target abstractive summary to calculate the reward.

\subsection{Content Planning}
We define the content plan as an ordered sequence of event-based representations. Each event is represented as a tuple containing a predicate (typically a verb or verb compound that describes the main action) and its primary arguments (e.g., subject, object, indirect object, and prepositional object). These tuples are typically structured as subject-verb-object triples. The content plan generator module utilizes the salient sentences from the previous module to construct a sequence of event tuples.

To train this module, both salient content and content plans for each document are required as input and output, respectively. However, since summarization corpora consist only of complete documents and their corresponding abstractive summaries, we need to extract sentence-level binary labels, which can be concatenated to form the salient content used as input for the plan generator module. Following \citet{liu2019extracting}, we select sentences for the summary based on their ROUGE-2 similarity score with the entire gold summary. This subset of sentences serves as the salient content for training our plan generator. During training, we utilize oracle-extracted salient sentences as input. In contrast, during inference, we rely on model-extracted salient sentences, which may be of lower quality and potentially introduce exposure bias in the plan generator. To mitigate this issue, we implement a hybrid training setup that combines outputs from the previous module with the golden extracted sentences, which enhances the model's robustness against noisy extractions during inference.

To obtain the content plan from golden summary, we utilize Plumber \cite{jaradeh2021better}, a tool designed for creating knowledge graph triples from unstructured text. Plumber dynamically generates suitable information extraction pipelines, offering a total of 264 distinct pipelines. It employs a RoBERTa-based classification model to extract contextual embeddings from the input and identify an appropriate pipeline, which includes components for coreference resolution, triple extraction, entity linking, and relation linking to knowledge graphs. For our purposes, we focus primarily on coreference resolution and triplet extraction to derive the event representations for each sentence in the summary. These event representations are then concatenated in the order of the summary sentences to form the golden content plan.

Our plan generator employs an Unlimiformer model \cite{bertsch2024unlimiformer}, a retrieval-based approach to augment pretrained language models to accept inputs of unbounded length at test time, to deal with longer inputs.  Unlimiformer uses BART-encoder \cite{lewis2020bart} to encode overlapping chunks of the input and keep only the middle half of the encoded vectors from each chunk, to ensure that the encodings have sufficient context on both sides. Then it constructs a k-nearest-neighbor index over the hidden states of all input tokens. Then, every standard cross-attention head in every decoder layer queries the kNN index, such that the kNN distances are the attention dot-product scores, and attends only to the top-k input tokens. We train this model with our curated input-output sequences. 

\input{text/tab_data_stat}

\subsection{Summary Generation} 
This module takes the salient content selected by the content selection module and the generated event-based plan from the planning module to produce the final summary, ensuring coherence and adherence to the event-based narrative structure. We utilize oracle-extracted salient content and content plans derived from the target summary, as described in previous modules, to construct the input for training our summary generator. To address the exposure bias problem, we implement hybrid training, incorporating some inputs populated by model-extracted content and model-generated content plans, which enhances the model's robustness during inference. Our summary generator employs a Longformer Encoder-Decoder (LED) \cite{beltagy2020longformer} model, which employs local-global attention mechanism in the encoder to effectively handle longer input lengths.

%% file: text/tab_data_stat.tex
\begin{table}[]
\setlength{\tabcolsep}{3pt} 
\scalebox{0.75}{
\begin{tabular}{lrrrrrrr}
\toprule
            & \multicolumn{2}{c}{\textbf{Doc}}   & \multicolumn{2}{c}{\textbf{Sum}}  & \textbf{Train} & \textbf{Valid} & \textbf{Test} \\
\cmidrule(lr){2-3} \cmidrule(lr){4-5}
            & \textbf{Tok.} & \textbf{Sent.}          & \textbf{Tok.} & \textbf{Sen.}        &                 &                &               \\ 
\midrule
MLS-Long    & 103712                     & 3150                      & 726                        & 24                        & 3177                      & 454                       & 908                      \\
MLS-Short   & 137042                     & 4170                      & 143                        & 5                         & 2210                      & 312                       & 616                      \\
CanLII      & 4914                       & 174                       & 291                        & 11                        & 839                      & 104                      & 106                      \\
SuperSCOTUS & 5406                       & 165                       & 902                        & 29                        & 3246                      & 406                       & 406      \\ \bottomrule                
\end{tabular}}
\caption{Statistics of datasets used in this paper.}
\label{tab:data_stat}
\end{table}

%% file: text/experiments.tex
\textbf{Datasets} We experiment with the following four legal case summarization datasets: 

\input{text/tab_results_main}

\noindent \emph{MultiLexSum-Long/Short} \cite{shen2022multi} comprises expert-written summaries for U.S. federal civil rights lawsuit cases, derived from multiple source documents such as complaints, settlement agreements, opinions, and orders. It features two summary granularities: (a) Long (L) summaries that detail the case background, parties, and events, and (b) Short (S) summaries that provide a concise overview in a single paragraph.

\noindent \emph{CanLII} \cite{xu2021accounting} dataset includes legal case decisions and human-prepared summaries sourced from the Canadian Legal Information Institute, covering a variety of legal claims and issues presented in Canadian courts
\input{text/tab_analysis_coperlex}

\noindent \emph{SuperSCOTUS} \cite{fang2023super} is a multi-source dataset of U.S. Supreme Court cases, integrating resources such as oral arguments, post-hearing annotations, and summaries, including case Opinions. For our work, we utilize the summarization dataset, which generates the Syllabus from the Court’s majority opinions excluding concurring and dissenting opinions. The Syllabus, prepared by the Reporter of Decisions—a statutory officer under the Court’s direction— summarizes the case background, details from lower courts, and the court's reasoning.  

Detailed statistics for these datasets are presented in Table \ref{tab:data_stat}. The MLS-Long and Short datasets feature longer input lengths, highlighting their multi-document nature. Both MLS-Long and SuperSCOTUS exhibit longer summary lengths, while MLS-Short presents a greater challenge due to its higher compression ratio.

\noindent \textbf{Metrics:} We evaluate the generated summaries using ROUGE-1, 2, and L \cite{lin2004rouge} to assess lexical overlap with reference paragraphs, and BERTScore \cite{zhang2019bertscore} for semantic similarity between the generated and reference summaries. For faithfulness, we use AlignScore \cite{zha2023alignscore} to measure factual consistency through a unified alignment function between the input context and generated text. Additionally, coherence and fluency scores are reported using the UniEval metric \cite{zhong2022towards}.

\noindent \textbf{Baselines:} We compare our CoPERLex approach with several abstractive summarization models designed for long documents: (i) LED \cite{beltagy2020longformer} (Longformer Encoder-Decoder), (ii) PRIMERA \cite{xiao2022primera}, (iii) LongT5 \cite{guo2022longt5}, (iv) SLED \cite{ivgi2023efficient}, and (v) Unlimiformer \cite{bertsch2023unlimiformer}. Detailed descriptions of the baselines and implementation specifics for CoPERLex can be found in App. \ref{app:baselines} and \ref{app:impl_details} respectively.

\subsection{Results}
From Table \ref{tab:main_results}, we observe that LED consistently outperforms PRIMERA across all datasets, which can be attributed to the larger input length of LED (16k vs. 4k). This highlights the absence of lead bias in legal datasets, unlike news datasets, and underscores the need for better long-context modeling strategies. LongT5 also surpasses PRIMERA, likely due to its extended input length, but still falls short of LED's performance.

SLED performs comparably to these long-range models on MLS-Long, MLS-Short, and SuperSCOTUS, while outperforming them on CanLII. Unlimiformer generally outperforms all the long-range models except on SuperSCOTUS, demonstrating that using off-the-shelf, short-range pre-trained models like BART within the SLED or Unlimiformer framework yields competitive results without the need for long-range pre-training or specialized summarization objectives. Additionally, Unlimiformer's strategy of attending to the top-k input keys proves to be an effective approximation of full attention. However, while SLED and Unlimiformer show competitive performance in content-based metrics like ROUGE, BERTScore, and AlignScore, they underperform in coherence and fluency across most datasets, except for CanLII in terms of coherence. This suggests that approximate attention and short-range models affect stylistic elements, indicating the need for improved approaches to enhance these aspects.

Overall, our proposed CoPERLex consistently outperforms all tested models across all datasets in most of the content and style metrics. 
This can be attributed to its modular framework handling three complementary steps of content selection to filter irrelevant information, event-centric outline generation to structure the narrative and ensure coherence and final synthesis to combine the outline with selected content into a cohesive summary. This division prevents the end-to-end model from becoming overwhelmed, allowing it to produce more concise, coherent, and contextually accurate outputs.
The improvements are higher in MLS-Long and SuperSCOTUS, which feature longer summaries (Table \ref{tab:data_stat}), particularly evident in coherence, fluency, and AlignScore, as end-to-end models struggle to balance factual accuracy with structural integrity. CoPERLex, by managing sub-tasks independently, performs each step more effectively. For datasets with shorter summaries, such as CanLII and MLS-Short, end-to-end models are comparable or even superior in style-based metrics. However, CoPERLex excels in factual accuracy, as it better handles the lengthy nature of inputs with content selection, where end-to-end models often fail to preserve source fidelity,  demonstrating the advantage of multi-step approach.

\noindent \textbf{Ablation Study} To assess the contribution of each component in CoPERLex, we conducted an ablation study by testing four variations: (a) without the content selection component (w/o Con. Sel.), where we use the entire document to produce a plan and generate the summary, bypassing content filtering; (b) without the planning component, where we directly extract salient content and generate the summary in a traditional extractive-then-abstractive approach; (c) without both the content selection and planning components, using the entire document to produce a summary without any intermediate steps and (d) without Hybrid Training in both the plan generation and summary generaiton modules. The results for these variants, evaluated on the MLS-Long and CanLII datasets, are presented in Table \ref{tab:main_coperlex}. Our findings reveal that removing both components leads to the most significant performance degradation. For MLS-Long, the removal of content selection had the greatest impact, as the longer input made it more challenging for the planning model to both understand salient content and generate a coherent summary structure. In contrast, for CanLII, removing the planning component was more detrimental, as the shorter input meant the absence of content selection had less of an effect.  The lack of a planning results in disjointed summaries that fail to reflect the logical progression of legal arguments or case facts, evidenced by coherence, align scores. 
Finally, removing the hybrid training mechanism degrades CoPERLex’s performance, emphasizing the robustness offered by this.

\noindent \textbf{Plan Representation} We evaluated various representations for the planning component: (a) entity-centric plans, represented by entity chains—ordered sequences of entities mentioned in the summary, as proposed by \citet{narayan2021planning}; (b) event-centric plans, where events are represented as tuples of subject, verb, and object, extracted through dependency parsing to capture syntactic relationships. This approach, used in prior work by \citet{joshi2023u} for legal case retrieval, is adapted here for summarization; (c) event-centric plans from Plumber \cite{jaradeh2021better}, a framework that uses a multi-pipeline approach to extract entities, relations, and handle coreferences, which we adopt in CoPERLex. We provided the summarization model with oracle plans (extracted from the reference summaries) along with the source documents, and reported the results on the MLS-Long dataset in Table \ref{tab:main_plan}. Our findings show that event-centric plans consistently outperform entity-centric ones, reflecting the unique narrative structure of legal documents. While a limited number of entities (such as the involved parties) remain central, the complexity of the events surrounding them intensifies over time. This growing complexity is essential for capturing the evolving arguments, decisions, and legal reasoning in a case. Event-centric representations are thus more effective for summarizing legal documents, as they focus on dynamic interactions among entities rather than static entities. Among event-centric methods, Plumber's orchestration of pipelines for entity and relation extraction delivers superior representations, allowing for a more accurate and coherent depiction of legal discourse, resulting in improved performance.
\input{text/tab_analysis_plan}

\input{text/tab_analysis_content}
\noindent \textbf{Content Selection} We analyze various extractive algorithms for the content selection component, comparing MemSum with BERT-Ext \cite{miller2019leveraging} and HiBERT \cite{zhang2019hibert}. BERT-Ext is an unsupervised approach that embeds source documents and selects  k key sentences from the embedding clusters, while HiBERT utilizes a hierarchical document encoder pre-trained on longer contexts using unlabeled data. We report the performance of the extractive summary against the abstractive reference summary using ROUGE scores in Table \ref{tab:main_content} on MLS-Long dataset. The superior performance of MemSum can be attributed to two main factors: (1) its stopping criterion is modeled as an action in RL framework that reduces the necessity to predefine the number of required sentences, and (2) its effective use of extraction history, allowing for iterative sentence selection rather than relying on a one-step extraction process.

\input{text/tab_analysis_arch}
\noindent \textbf{Plan Generation} In the plan generation phase, we utilize the entire document as input for the planning model to produce event-based tuples as outlines. We evaluate the performance of the LED model and the Unlimiformer model by measuring ROUGE scores between the generated plans and the extracted plans from the reference summary. As shown in Table \ref{tab:main_arch}, Unlimiformer consistently outperforms LED, particularly for longer sequences in R-2 and R-L. Consequently, we chose  Unlimiformer as our plan generator in CoPERLex.

\noindent \textbf{Summary Generation} For the summary generation, we use oracle plans extracted from the reference summary as input to the summarizer model, alongside the document. Again, we explore the performance of both LED and Unlimiformer models. Table \ref{tab:main_arch} indicates that LED significantly outperforms Unlimiformer in this context, which contrasts with our earlier findings in Table \ref{tab:main_results} on the Plan Generation section. We attribute this discrepancy to Unlimiformer's kNN-based attention mechanism, which may struggle to adhere to the provided plan triples due to its lack of an exclusive focus mechanism. This limitation hinders its ability to concentrate on planning tuples while retrieving k tokens during decoding. Therefore, we opted to use LED as our summary generator in CoPERLex.

\noindent \textbf{Case Study:} In this section, we explore two key advantages of our event-based content planning framework: (i) how it enhances structural coherence and faithfulness to the source document and (ii) how it offers controllability for generating more customized summaries. These aspects are illustrated with examples in App. \ref{app:case_study}. 

Without a content plan, summaries often lack structure and coherence, leading to irrelevant or incomplete information. They may even fabricate details that are not aligned with the source document. As seen in Table \ref{tab:case1}, key legal actions, such as the filing of a lawsuit, are omitted, resulting in incomplete summaries that fail to capture critical events. The date of the filing is fabricated as the date of settling the lawsuit, creating unfaithful summaries. Entity-based planning offers some improvement by grounding the summary in entities or crucial dates, which reduces the likelihood of fabrication. However, as observed, focusing solely on entities can still miss essential actions and relationships, leading to unreliable summaries, such as misinterpreting the filing date as the intervention date. This occurs because events often revolve around entities, and without a clear grounding strategy, multiple possibilities arise for interpreting those entities, making it challenging to create a faithful summary. Our event-based planning approach addresses these challenges by grounding the summary in specific actions and interactions, rather than just the entities, making the summary more anchored to the source document.

In Table \ref{tab:case2}, we modified the event sequence by incorporating relevant details (such as the `requirement for signatures from 5\% of registered voters') and removing irrelevant elements (like `seeking relief'). This adjustment affected the final summary, showcasing how users can customize content for improved conciseness and informativeness through event-based planning. Additionally, this process illustrates a potential method for enhancing summary faithfulness by employing similar strategies to develop automated methods for verifying and refining unfaithful event tuples, leading to more reliable outputs.

%% file: text/tab_results_main.tex
\begin{table*}[!t]
\center
\begin{tabular}{ll ccccccc}
\toprule
                             &  &    \textbf{R-1} & \textbf{R-2} & \textbf{R-L} &  \textbf{BS}    & \textbf{AS}    & \textbf{Coh}   & \textbf{Flu}   \\ \midrule
\multirow{6}{*}{MLS-Long}
                             & LED           & {47.83}        & {24.95}        & {28.69}        & {86.31}       & {64.22}       & {75.50}       & {78.79}     \\  
                             & PRIMERA        & {45.14}        & {23.68}        & {27.47}        & {85.46}       & {62.48}       & {75.83}       & {78.23}     \\  
                             & LongT5          & {46.26}        & {24.22}        & {27.76}        & {85.82}       & {63.73}       & {75.79}       & {78.78}     \\  
                             & SLED-T5         & {46.73}        & {24.04}        & {28.24}        & {86.34}       & {65.18}       & {75.12}       & {77.66}     \\  
                             & Unlimiformer-T5 & {48.42}        & {24.56}        & {28.11}        & {86.18}       & {66.24}       & {74.82}       & {77.53}     \\  
                             & CoPERLEX        & $\textbf{50.24}^*$ & $\textbf{26.57}^*$ & $\textbf{29.24}^*$ & $\textbf{86.92}^*$ & $\textbf{67.51}^*$ & $\textbf{77.84}^*$ & $\textbf{80.21}^*$ \\ \midrule

\multirow{6}{*}{MLS-Short} 
                             & LED             & 45.66 & 22.46 & 30.64 & 87.94 & 61.45 & 68.15 & 81.46 \\  
                             & PRIMERA         & 42.27 & 21.68 & 29.18 & 86.93 & 61.57 & \textbf{69.40} & \textbf{81.88} \\  
                             & LongT5          & 44.24 & 22.04 & 29.87 & 87.68 & 61.25 & 69.21 & 81.39 \\  
                             & SLED-T5         & 45.52 & 22.27 & 30.12 & 88.12 & 61.87 & 68.03 & 80.65 \\  
                             & Unlimiformer-T5 & 45.46 & \textbf{23.13} & 30.77 & 88.05 & 62.51 & 67.64 & 80.76 \\  
                             & CoPERLEX        & $\textbf{46.18}^*$ & 23.04 & $\textbf{31.13}^*$ & $\textbf{88.69}^*$ & $\textbf{63.49}^*$ & 69.27 & 81.79 \\ \midrule

\multirow{6}{*}{CanLII} 
                             & LED             & 47.99 & 24.47 & 43.66 & 86.30 & 63.13 & 64.70 & 86.25 \\  
                             & PRIMERA         & 47.65 & 24.19 & 44.04 & 86.77 & 64.72 & 64.24 & 86.18 \\  
                             & LongT5          & 48.08 & 24.76 & 44.48 & 86.26 & 64.62 & 64.78 & 85.78 \\  
                             & SLED-T5         & 48.12 & 25.05 & 44.62 & 86.81 & 65.31 & 65.06 & 86.09 \\  
                             & Unlimiformer-T5 & 48.62 & 
                             25.62 & 44.58 & $\textbf{87.66}^*$ & 65.06 & 64.92 & 86.14 \\  
                             & CoPERLEX        & \textbf{49.19} & \textbf{25.88} & $\textbf{45.44}^*$ & 87.25 & $\textbf{66.23}^*$ & $\textbf{65.42}^*$ & $\textbf{86.53}$ \\ 
 \midrule
\multirow{6}{*}{SuperSCOTUS} 
                             & LED             & 51.14 & 24.82 & 28.11 & 82.12 & 56.70 & 63.50 & 67.60 \\  
                             & PRIMERA         & 48.65 & 24.17 & 26.69 & 82.88 & 56.25 & 63.77 & 67.12 \\  
                             & LongT5          & 49.28 & 24.88 & 27.37 & 83.34 & 56.39 & 64.16 & 67.30 \\  
                             & SLED-T5         & 50.36 & 24.61 & 28.24 & 82.91 & 56.18 & 63.12 & 66.18 \\  
                             & Unlimiformer-T5 & 50.21 & 24.49 & 27.79 & 83.18 & 57.12 & 63.29 & 66.49 \\  
                             & CoPERLEX        & $\textbf{52.28}^*$ & $\textbf{25.78}^*$ & $\textbf{28.94}^*$ & $\textbf{83.52}^*$ & $\textbf{59.38}^*$ & $\textbf{66.21}^*$ & $\textbf{69.49}^*$ \\ 
 \bottomrule
\end{tabular}
\caption{Performance comparison of CoPERLex across four datasets. BS, AS, Coh, and Flu represent BERTScore, AlignScore, Coherence and Fluency, respectively. Entries marked with * indicate statistically significant improvement over the second-best performance, using the Wilcoxon signed-rank test with a 95\% confidence interval.}
\label{tab:main_results}
\end{table*}

%% file: text/tab_analysis_coperlex.tex
\begin{table*}[]
\setlength{\tabcolsep}{3pt} 
\centering
\scalebox{0.88}{
\begin{tabular}{lccccccccccccccc}
\toprule
                       & \multicolumn{7}{c}{\textbf{MLS-Long}}                          & \multicolumn{7}{c}{\textbf{CanLII}}                            \\ 
\cmidrule(lr){2-8} \cmidrule(lr){9-15}
                       & \textbf{R-1}   & \textbf{R-2}   & \textbf{R-L}   & \textbf{BS}    & \textbf{AS}    & \textbf{Coh}   & \textbf{Flu}   & \textbf{R-1}   & \textbf{R-2}   & \textbf{R-L}   & \textbf{BS}    & \textbf{AS}    & \textbf{Coh}   & \textbf{Flu}   \\ 
\midrule
CoPERLex            & \textbf{50.24} & \textbf{26.57} & \textbf{29.24} & \textbf{86.92} & \textbf{67.51} & \textbf{77.84} & \textbf{80.21} & \textbf{49.19} & \textbf{25.88} & \textbf{45.44} & \textbf{87.25} & \textbf{66.23} & 65.42 & \textbf{86.53} \\  
w/o Content Sel.    & 48.14 & 25.14 & 28.54 & 86.42 & 66.03 & 75.34 & 79.78 & 49.06 & 25.41 & 45.28 & 86.72 & 66.14 & \textbf{65.72} & 86.18 \\  
w/o Planning        & 49.98 & 25.22 & 28.91 & 86.44 & 66.64 & 75.84 & 79.27 & 48.17 & 24.22 & 44.09 & 86.37 & 64.36 & 65.05 & 86.27 \\  
w/o CS \& Planning  & 47.83 & 24.95 & 28.69 & 86.31 & 64.22 & 75.50 & 78.79 & 47.99 & 24.47 & 43.66 & 86.30 & 63.13 & 64.70 & 86.25 \\  
w/o Hybrid Train. & 48.61 & 25.36 & 28.66 & 86.45 & 63.87 & 76.12 & 79.16 & 48.38 & 25.03 & 44.81 & 86.54 & 63.29 & 65.12 & 86.18 \\ 
 \bottomrule
\end{tabular}}
\caption{Ablation study to assess effect of different components in CoPERLex framework. Con. Sel. (CS) \& Pl. denote Content Selection \& Planning components respectively.}
\label{tab:main_coperlex}
\end{table*}

%% file: text/tab_analysis_plan.tex
\begin{table}[]
\setlength{\tabcolsep}{3pt} 
\scalebox{0.9}{
\begin{tabular}{ll ccccccc}
\toprule
 & \textbf{R-1}   & \textbf{R-2}   & \textbf{R-L}   & \textbf{BS}    & \textbf{AS}  & \textbf{Coh}      \\ \midrule
Entities             & 54.14        & 30.16        & 38.56        & 88.66       & 77.13       & 80.45        \\  
UCREAT               & 57.72        & 32.61        & 41.91        & 89.61       & 78.46       & 80.65        \\  
Plumber              & \textbf{59.65} & \textbf{36.16} & \textbf{44.26} & \textbf{89.72} & \textbf{80.71} & \textbf{82.68} \\

 \bottomrule
\end{tabular}}
\caption{Comparison of various planning approaches on the MLS-Long dataset, using oracle plans derived from the reference summary as input to the summary generation module.}
\label{tab:main_plan}
\end{table}

%% file: text/tab_analysis_content.tex
\begin{table}[ht]
\centering
\setlength{\tabcolsep}{3pt} 
\scalebox{0.9}{
\begin{tabular}{lcccccc}
\toprule
\textbf{} & \multicolumn{3}{c}{\textbf{MLS-Long}} & \multicolumn{3}{c}{\textbf{CanLII}} \\ 
\cmidrule(lr){2-4} \cmidrule(lr){5-7} 
         & \textbf{R-1} & \textbf{R-2} & \textbf{R-L} & \textbf{R-1} & \textbf{R-2} & \textbf{R-L} \\ 
\midrule
BERT-Ext             & 34.63        & 9.10         & 13.17        & 38.43       & 23.75       & 20.56        \\  
HiBERT               & 36.28        & \textbf{11.28} & 14.83        & 41.27       & 24.74       & 26.75        \\  
MemSum               & \textbf{36.92} & \textbf{12.85} & \textbf{15.56} & \textbf{42.16} & \textbf{25.19} & \textbf{28.16} \\ 
\bottomrule
\end{tabular}}
\caption{Comparison of different extractive approaches for Content Selection.}
\label{tab:main_content}
\end{table}

%% file: text/tab_analysis_arch.tex
\begin{table}[]
\setlength{\tabcolsep}{3pt} 
\centering
\scalebox{0.87}{
\begin{tabular}{lcccccc}
\toprule
                       & \multicolumn{3}{c}{\textbf{Plan Gen.}} & \multicolumn{3}{c}{\textbf{Sum. Gen. with OP}} \\
\cmidrule(lr){2-4} \cmidrule(lr){5-7}
                       & \textbf{R-1} & \textbf{R-2} & \textbf{R-L} & \textbf{R-1} & \textbf{R-2} & \textbf{R-L} \\ 
\midrule
LED            & \textbf{35.18} & 15.03        & 20.81              & \textbf{59.65} & \textbf{36.16} & \textbf{44.26} \\  
Unlimiformer      & 34.96        & \textbf{15.65} & \textbf{21.44}    & 57.16       & 33.88       & 41.18        \\ 
\bottomrule        
\end{tabular}}
\caption{Comparison of model architectures for Plan Generation and Summary Generation using Oracle Plans (OP) on MLS-Long dataset.}

\label{tab:main_arch}
\end{table}

%% file: text/conclusion.tex
In this paper, we introduced CoPERLex, a novel approach for legal case summarization that integrates salient content selection and content planning with event-centric representations. Our experiments showed that utilizing structured event-based plans significantly enhances coherence and narrative flow, outperforming state-of-the-art models across four legal datasets. By capturing relationships between entities within legal texts, CoPERLex enhances summarization quality and paves the way for future research in legal NLP, underscoring the importance of event-driven frameworks in addressing the complexity of legal narratives.

%% file: text/limitations.tex
While CoPERLex employs event-centric representations to enhance narrative flow in legal case summarization, its current approach oversimplifies the representation of events and their interrelationships. Legal cases often involve intricate temporal relations, event co-references, and complex cause-and-effect chains, necessitating more sophisticated narrative modeling. Future work could improve CoPERLex by incorporating advanced representations that effectively capture these temporal dependencies, as well as utilizing graph-based structures to represent the interplay among multiple events and legal arguments. This enhancement would enable the system to better address more complex legal cases, where the sequence and interaction of events are pivotal to case outcomes.

Another limitation of this work lies in the evaluation methodology. While CoPERLex’s performance is assessed using widely established metrics such as ROUGE, BERTScore, and AlignScore, these metrics are limited in their ability to fully capture the unique complexities of legal texts, such as legal reasoning and the argumentative structure. Future research could focus on developing domain-specific evaluation metrics tailored to the needs of legal summarization, potentially involving more qualitative measures of reasoning accuracy and argument structure. Additionally, the lack of validation by legal experts is a notable limitation of this study. Direct feedback from legal professionals could provide invaluable insights into the practical utility of the summaries generated by CoPERLex, but this was not feasible due to the unavailability of legal experts for evaluation.

Finally, CoPERLex has primarily been evaluated on English-language legal datasets, which limits its applicability to other languages. Extending the model to non-English legal documents presents an interesting avenue for future exploration.  Moreover, its evaluation has been conducted in a same-jurisdiction setting—i.e., trained and tested on cases from the same jurisdiction. Given the diverse nature of legal systems across countries and the limited availability of annotated summarization datasets, ensuring cross-jurisdictional generalizability of summarization systems becomes essential \cite{santosh2024beyond}. Investigating CoPERLex in cross-jurisdictional settings remains an another interesting future direction.

%% file: text/ethics.tex
The datasets utilized in this research were either publicly accessible or obtained through agreements with their respective owners, ensuring ethical compliance in their usage. While these datasets are not anonymized, they comprise legal documents that are already in the public domain or made available through official channels. As a result, their use in our study does not pose any additional harm beyond their existing availability.

The development of legal summarization systems like CoPERLex presents several ethical challenges. First, while automating the summarization of legal documents can significantly reduce the workload for legal professionals, it is crucial to recognize that such systems are not substitutes for human expertise, particularly in sensitive legal contexts. The summaries generated by CoPERLex may fail to capture all the nuances of complex legal cases, which could lead to misinterpretation or oversimplification of critical details. To address this concern, the model’s outputs should be regarded as assistive tools rather than authoritative sources and must be thoroughly reviewed by qualified legal professionals before any formal application.

Second, the biases inherent in the training data may influence the fairness and objectivity of the generated summaries. Legal datasets often reflect societal biases and historical injustices, raising the risk that CoPERLex could inadvertently perpetuate these biases in its outputs. Therefore, it is essential to implement careful selection of training data, continuous monitoring, and effective bias mitigation strategies to minimize these risks.

%% file: text/appendix.tex
\section{Baselines}
\label{app:baselines}
\noindent \textbf{LED} \cite{beltagy2020longformer} is a longformer variant equipped with both encoder and decoder. In the encoder, it uses efficient local+global attention pattern instead of full self-attention, while the decoder utilizes full quadratic attention. It is initialized from the pre-trained BART model \cite{lewis2020bart}. We use the LED-base version, which can handle input lengths of up to 16,384 tokens.

\noindent \textbf{PRIMERA} \cite{xiao2022primera} is initialized with LED model and pre-trained with a novel summarization-specific entity-based sentence masking objective, similar to Pegasus, but can be applied for longer inputs, motivated by Gap Sentence Generation objective of Pegasus \cite{zhang2020pegasus}. It can handle upto 4096 tokens. 

\noindent \textbf{LongT5} \cite{guo2022longt5} scales the input length by a new attention mechanism called as transient global attention which mimics local+global attention from ETC \cite{ainslie2020etc} and adopt summarization specific pre-training from PEGASUS into the T5 model \cite{raffel2020exploring} for longer sequences. We use the LongT5-base version which can handle upto 16384 tokens.

\noindent \textbf{SLED} \cite{ivgi2023efficient} processes long sequences by partitioning the input into overlapping chunks and encode each chunk with a short-range pre-trained models encoder. It relies on decoder to fuse information across chunks attending to all input tokens, similar to fusion-in-decoder \cite{izacard2021leveraging}. SLED can be applied on top of any short-range models and thus we derive SLED-BART-base by applying on BART-base \cite{lewis2020bart}. 

\noindent \textbf{Unlimiformer} \cite{bertsch2023unlimiformer} adopts a strategy similar to SLED, but rather than attending to all input tokens, it focuses exclusively on the top-k tokens retrieved from a k-nearest-neighbor index constructed over the hidden states of all input tokens at every decoder layer. This helps to process unbounded length during testing, a key differentiator from SLED, which is constrained by memory when attending to all input tokens in the decoder. We derive Unlimiformer-BART-base model.

\section{Implementation Details}
\label{app:impl_details}
For the content selection selection, we created high-ROUGE episodes for the training set as described in original paper of \cite{gu2022memsum} with a branching size of 2. We select the highest-scoring model based on the validation ROUGE-L scores. During inference, we set the stopping threshold to 0.6 and the maximum extracted sentences as 45 sentences. We train the model end-to-end for 20 epochs with a batch size of 1 using Adam Optimizer \cite{kingma2014adam} with $beta_1$ and $beta_2$ set to 0.9 and 0.999.  We choose learning rate to be 1e-4. During training, local sentence embeddings are kept fixed using pretrained Glove word embeddings of dimenssion d = 200 \cite{pennington2014glove}. We set number of Bi-LSTM layers in local sentence and  and global context encoder to 2. For, Extraction History Encoder, we use 3 attention layers with 8 attention heads and a feed-forward hidden layer of a dimension of 1024 and dropout rate of 0.1. We run the Plumber framework on the summaries of all datasets to obtain the golden plans and prepare the plan generation datasets. We trained the plan generation unlimformer models for 10 epochs on a batch size of 1. We set the maximum input length to 16384 tokens and the maximum generation length to 512 tokens. We set the chunk size and chunk overlap to 256 and 0.5 respectively. The model is trained end-to-end with learning rate of 0.0001 with Adam optimizer. Best model is chosen based on validation score of ROUGE-L between generated and golden plan. For summarization model, we trained LED for 15 epochs on a batch size of 1 and with a learning rate of 3e-5 using Adam Optimizer. The maximum input length is set to 16384 tokens and the maximum output length to 1024 tokens and the attention window size is set to 256 tokens. We chose the best model based on validation ROUGE-L score. 

\input{text/tab_case_2}

\section{Case Study}
\label{app:case_study}
Table \ref{tab:case1} illustrates a summary generated using a baseline model without planning, alongside models that incorporate entity and event-centric planning. This case study demonstrates how event-centric planning can enhance both the structural coherence and the faithfulness of a summary to the source document. 
\input{text/tab_case_1}

Table \ref{tab:case2} showcases the customization process of modifying plans, which can influence the final summary. This case study illustrates how event-centric planning offers users the flexibility to create tailored summaries.

%% file: text/tab_case_2.tex
\begin{table}[!htb]
\setlength{\tabcolsep}{3pt} 
\centering

\begin{tabular}{p{8cm}}
\toprule
\emph{Plan A:} \\
 Two third-party nominees - filed
- case | case - filed - in light of public health risks of COVID-19 | requiring signatures - violated - the First and Fourteenth Amendments | Court - ordered - defendants to reduce signature requirements | plaintiffs - sought -relief | defendants - reduce - signatures requirements by 30\% \\

\emph{Summary A:} \\
Two third-party nominees for the Georgia State House of Representatives filed this case in light of public health risks of COVID-19,  alleging that the requirement to gather more than 20,000 signatures to qualify for the ballot violated their First and Fourteenth Amendment rights. The plaintiffs sought declaratory and injunctive relief to which the court ordered the defendants to reduce the required signatures by 30\% \ldots \\
\midrule
\emph{Plan B:} \\
 Two third-party nominees - filed
- case | lawsuit - filed - in light of public health risks of COVID-19 | \textcolor{blue}{signatures - required from- 5\% of the registered voters} | requiring signatures - violated - the First and Fourteenth Amendments | Court - ordered - defendants to reduce signature requirements | \textcolor{blue}{\st{plaintiffs sought relief}}|
defendants - reduce - signatures requirements by 30\% \\

\emph{Summary B:} \\
Two third-party nominees for the Georgia State House of Representatives filed this case in light of public health risks of COVID-19,  alleging that the requirement to \textcolor{blue}{signatures from 5\% of the registered voters in the district where they are running,} to qualify for the ballot violated their First and Fourteenth Amendment rights. The court ordered the defendants to reduce the required signatures by 30\% \ldots \\
\bottomrule        
\end{tabular}
\caption{Case Study illustrating how plans offer controllability to generate customised summaries. }
\label{tab:case2}
\end{table}

%% file: text/tab_case_1.tex
\begin{table*}[]
\setlength{\tabcolsep}{3pt} 
\centering

\begin{tabular}{p{15cm}}
\toprule
\textbf{Reference Summary} \\
\midrule
\ldots The Defendants were Furniture Distributors, Inc., Kimbrell's of Statesville, NC, and Kimbrell's, Inc., all
associated as a furniture store. The controversy involved the alleged sexual harassment and constructive
discharge of an employee in violation of Title VII of the Civil Rights Act of 1964. The lawsuit was filed in
August of 2002 and was eventually resolved by consent decree in October of 2003. The complainant, who
intervened in the case, was awarded \$25,000 and a guarantee of neutral references. \ldots \\
\midrule
\midrule
\textbf{w/o Planning} \\
\midrule
\ldots against Furniture Distributors, Inc., Kimbrell's Furniture, Inc., and Kimbrell's, Inc. in the U.S. District
Court for the Western District of North Carolina alleging discrimination on the basis of sex in violation of
Title VII of the Civil Rights Act of 1964. Specifically, the complaint alleged that the defendant subjected
the charging party, a female employee, to a sexually hostile work environment and constructively
discharged her. Following some discovery, \textcolor{red}{the parties settled the lawsuit in August 2002 through a
consent decree}. The three-year decree, containing non-discrimination and non-retaliation clauses,
required the defendant to: report on complaints and make other compliance reports, allow the EEOC
access for monitoring \ldots \\
\midrule
\midrule
\textbf{Planning with Events} \\
\midrule
\emph{Plan:} \\
\ldots Furniture Distributors, Inc. | Kimbrell's Furniture | Statesville | NC | Kimbrell's ||| August of 2002 | October of 2003 ||| 25,000 ||| \ldots \\
\emph{Summary:} \\
\ldots The defendant was Furniture Distributors, Inc., a furniture manufacturer, doing business as Kimbrell's Furniture, a general store in Statesville, NC, and Kimbrell's, Incorporated collectively doing business as Kimbrell's. The complaint alleged that the defendant subjected the charging party to a sexually hostile work environment and constructively discharged her. \textcolor{red}{The charging party intervened in the lawsuit in August of 2002} and in October of 2003 the parties settled. The consent decree required the
defendant to pay the charging party \$25,000.  \ldots \\
\midrule
\midrule
\textbf{CoPERLex: Planning with Events} \\
\midrule
\emph{Plan:} \\
\ldots  Defendants - were -
Furniture Distributors | controversy - involved - sexual harassment of employee | sexual harassment - violation - Title VII of the Civil Rights Act of 1964 | Lawsuit - filed - August 2002 | Lawsuit - resolved - consent decree | Resolution - occurred - October 2003 | Complainant- awarded - \$25,000 \ldots \\
\emph{Summary:} \\
\ldots The Defendants were Furniture Distributors, Inc., Kimbrell's Furniture, Inc., and Kimbrell's, Inc., all
associated as a furniture store. The controversy involved sexual harassment of an employee in violation
of Title VII of the Civil Rights Act of 1964. The lawsuit was filed in August of 2002 and was eventually
resolved by consent decree in October of 2003. The complainant was awarded \$25,000.00. \ldots \\ 
\bottomrule  
\bottomrule  
\end{tabular}
\caption{Case study demonstrating the enhancement of structural coherence and faithfulness to source documents through event-centric planning.}
\label{tab:case1}
\end{table*}